%% file: ICVGIP-Latex-Template.tex
\documentclass[sigconf]{acmart}

\usepackage{booktabs} 

\usepackage{hyperref}  
\usepackage{hyperxmp}  
\setcopyright{rightsretained}

\acmConference[ICVGIP'24]{15th Indian Conference on Computer Vision, Graphics and Image Processing}{December 2024}{Bangalore, India}
\acmYear{2024}
\copyrightyear{2024}

\acmPrice{15.00}

\begin{document}
\title{Image Generation from Image Captioning - Invertible Approach}

\author{Nandakishore S Menon$^*$}
\affiliation{%
  \institution{IBM Research, Bangalore}
  \country{India}
}
\author{Chandramouli Kamanchi}
\affiliation{%
  \institution{IBM Research, Bangalore}
  \country{India}
}

\author{Raghuram Bharadwaj Diddigi}
\affiliation{%
  \institution{IIIT Bangalore}
  \country{India}
}

\thanks{$^*$Work done at IIIT Bangalore, India. E-mail: \url{nandakishore2001menon@gmail.com}}
\renewcommand{\shortauthors}{}

\begin{abstract}
Our work aims to build a model that performs dual tasks of image captioning and image generation while being trained on only one task. The central idea is to train an invertible model that learns a one-to-one mapping between the image and text embeddings. Once the invertible model is efficiently trained on one task, the image captioning, the same model can generate new images for a given text through the inversion process, with no additional training. This paper proposes a simple invertible neural network architecture for this problem and presents our current findings. 
\end{abstract}

%
%



\maketitle

\input{samplebody-conf}

\bibliographystyle{ACM-Reference-Format}
\bibliography{ICVGIP-Latex-Template}

\appendix




\end{document}

%% file: samplebody-conf.tex
\section{Introduction}
We consider the problem of training a model on one task with the objective that the inversion task can be inferred with no additional training. Specifically, we train a neural network to approximate an invertible function that maps images to its text annotation. Subsequently, one can invoke the inverse process and generate images from the given text if the network is well-trained. There are many popular models for image generation in the literature, for example, using VAE in \cite{razavi2019generating}, GANs in \cite{bao2017cvae}, diffusion models \cite{ho2020denoising}. Moreover, \cite{ramesh2021zero,ramesh2022hierarchical} are popular models for generating images from text. However, most of these models are complex and computationally intensive.
Our goal is to build a simple model that is trained in image captioning and can be used directly for image generation. 

\section{Problem Formulation}
Let $I$ and $E_{\mathcal{I}}(.)$ denote an image and image encoder which provides vector representation (embedding) of the image and let $T^I$ and $E_{\mathcal{T}}(.)$ represent the corresponding annotation and text encoder. Our objective is to train an invertible function $f:E_{\mathcal{I}}(I) \mapsto E_{\mathcal{T}}(T^I)$. As a consequence, given a text annotation $T$, we can invoke $f^{-1}(E_{\mathcal{T}}(T))$ to obtain embeddings of the corresponding image. Subsequently, the image can be generated by passing this embedding to an image decoder, i.e., $D_{\mathcal{I}}(f^{-1}(E_{\mathcal{T}}(T)))$, where $D_{\mathcal{I}}$ is the image decoder corresponding to the encoder $E_{\mathcal{I}}(I)$.
\section{Proposed Solution}
We divide this section into two parts. First, we propose constructing an invertible neural network for general tasks. In the second subsection, we incorporate it for image annotation and generation. 
\subsection{Invertible Neural Networks}
Consider a neural network $f(x)$ that learns a function mapping inputs $x$ to outputs $y$. Mathematically, we can represent $f$ as a composition of linear transformations and non-linear activations:
\[
f(x) =  \sigma_n (A_n(\sigma_{n-1} (A_{n-1} (\ldots \sigma_1(A_1 x + b_1) \ldots) + b_{n-1}) + b_n),
\]
where $A_i$ are the weight matrices, $b_i$ are the bias vectors, and $\sigma_i$ are the activation functions.

If each $A_i$ and $\sigma_i$ are invertible, then the neural network $f(x)$ is invertible. This means we can find a function $f^{-1}$ such that $f^{-1}(f(x)) = x$, where
\begin{align*}
&f^{-1}(x) = A^{-1}_1 (\sigma^{-1}_1 (A^{-1}_2 (\sigma^{-1}_2 (\ldots A^{-1}_{n-1} (\sigma^{-1}_{n-1} \\ &(A^{-1}_n (\sigma^{-1}_n (x) - b_n)) - b_{n-1}) \ldots ) - b_1))).
\end{align*}

The neural network will be invertible as long as the weight matrices in its linear layers and the activation functions used are invertible. We can choose invertible activation functions like Leaky ReLU, but the weight matrices are adjusted throughout the training process in order to learn the function $f$.
This poses a challenge as we need a definitive way to ensure that every weight matrix in the neural network is invertible, even after training. 

To ensure that a neural network is invertible, all matrices involved in the network must be \emph{non-singular matrices}. A matrix is invertible if and only if its determinant is non-zero. Therefore, to force the neural network to be invertible, all weight matrices and any other matrices used in the network must have non-zero determinants.
We devise a novel strategy to enforce invertibility in the structure of the neural network itself. 

Let \( L \) and \( U \) be lower triangular and upper triangular matrices, respectively, with \( n \times (n-1)/2 \) trainable parameters each. These parameters, denoted as \( (l_{ij}) \) and \( (u_{ij}) \), are utilized to instantiate \( L \) and \( U \). 

The lower triangular matrix \( L \) is constructed as follows:
\[
L_{ij} = 
\begin{cases}
l_{ij} & \text{if } i > j, \\
1 & \text{if } i = j, \\
0 & \text{if } i < j.
\end{cases}
\]

Similarly, the upper triangular matrix \( U \) is constructed as follows:
\[
U_{ij} = 
\begin{cases}
u_{ij} & \text{if } i < j, \\
k_{i} & \text{if } i = j, \\
0 & \text{if } i > j.
\end{cases}
\]

Here, \( k_{i} \) represents a non-zero constant.
Triangular matrices have a unique property: their determinant is equal to the product of their diagonal elements. We will be leveraging this property in our approach to enforce invertibility.

The final weight matrix \( W \) for a linear layer in our modified neural network is obtained by multiplying the lower triangular matrix \( L \) and the upper triangular matrix \( U \). Mathematically, this operation can be represented as:
\[
W_{n \times n} = LU.
\]

Since \( \det(AB) = \det(A) \times \det(B) \), the determinant of the weight matrix \( W \) can be expressed as:
\[
\det(W) = \det(LU) = \det(L) \times \det(U).
\]

As both \( L \) and \( U \) are triangular matrices with non-zero constant diagonal elements, their determinants are non-zero. Thus, the determinant of \( W \) is also non-zero, ensuring the invertibility of the neural network.

\begin{table}[htbp]
\centering
    \caption{Experimental Results for Various Functions}
    \label{tab:experiment_results}
\begin{tabular}{|c|c|c|}
\hline
\textbf{Function}    & \textbf{\begin{tabular}[c]{@{}c@{}}Error in \\ Learning\end{tabular}} & \textbf{\begin{tabular}[c]{@{}c@{}}Error in \\ Inversion\end{tabular}} \\ \hline
Sine Function        & $1e^{-5}$                                                             & $1e^{-4}$                                                              \\ \hline
Polynomial Function  & $1e^{-6}$                                                             & $1e^{-6}$                                                              \\ \hline
Exponential Function & $1e^{-4}$                                                             & $1e^{-4}$                                                              \\ \hline
\end{tabular}
\end{table}
Table 1 presents the errors in learning the target function and predicting its inverse for different types of functions. Despite the complexity of the functions, the invertible neural network achieved low errors in both learning and inversion tasks, demonstrating its effectiveness in learning invertible mappings.

\subsection{Application to Image Generation}
The objective is to construct a model architecture that, when trained on image captioning, will not only perform well on predicting caption embeddings but also generate an appropriate image for a given text when the model is inverted.

To train our model for the image captioning task, we will use \emph{Flickr30k} image captioning dataset \cite{dataset}, which contains a diverse set of 31,000 images and corresponding captions.

The model is comprised of three components:
\begin{enumerate}
    \item \textbf{A pre-trained image autoencoder \cite{sonderby2017continuous}:} A trained convolutional autoencoder's encoder is used to get a latent representation from an image that can be used to map to a caption. The autoencoder's decoder would be used in the inverse task to reconstruct the image.
    \item \textbf{A pre-trained sentence embedding model:} For obtaining a vector representation of the caption strings, a pre-trained sentence embedding model is used \cite{reimers2019sentence, conneau2017supervised}.
    \item \textbf{An invertible neural network:} A neural network that follows the construction described in Section 
    3.1 would be utilized.
\end{enumerate}

Note that the dimension of the latent representation (when flattened) of the autoencoder and the sentence embedding dimensions must be the same. This is done so that matrices within the neural network remain square, which is a requirement for invertible matrices.

  \begin{figure}[htbp]
      \centering
      \includegraphics[scale=0.2]{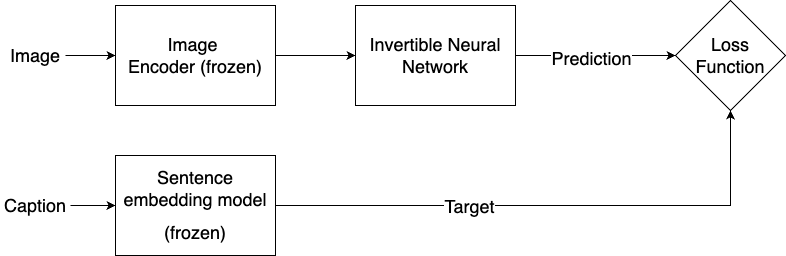}
      \caption{Architecture for training invertible neural networks to predict caption embeddings}
      \label{fig:arch1}
    \end{figure}
The image is passed into the frozen pre-trained convolutional autoencoder's encoder to produce the image embedding. The caption is passed through the frozen pre-trained Sentence Embedding model to produce the caption embedding, $y$. The image embedding is relayed to the invertible neural network to produce a prediction for the caption embedding, $y_{pred}$. MSE loss is computed between $y$ and $y_{pred}$, and the loss is backpropagated till the invertible neural network.




The inverse task can be understood better through Figure ~\ref{fig:arch1_inv} given below:
\begin{figure}[htbp]
      \centering
      \includegraphics[scale=0.2]{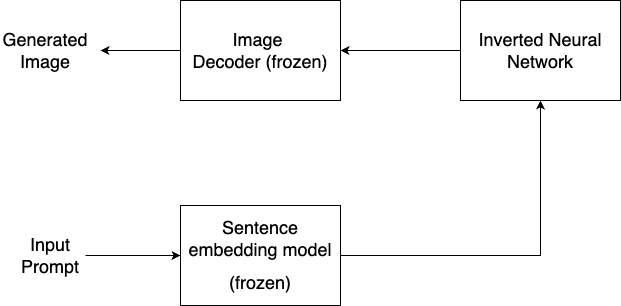}
      \caption{Flow of inverse task: Generating an image from a given caption}
      \label{fig:arch1_inv}
    \end{figure}
An input prompt, which describes the image we want to generate, is provided as input to the sentence embedding model. This sentence embedding is relayed to the inverted neural network. The predicted image embedding is passed through the pre-trained autoencoder's decoder to generate the image.


After training the model described in Figure ~\ref{fig:arch1}, we obtained the following results:
\begin{enumerate}
        \item \textbf{The model performs well on the task of image captioning}. Let $x$ be the image passed, $y_{pred}$ the predicted caption, and $\hat{y}$ be the target caption embedding.
        The error between $y_{pred}$ and $\hat{y}$ was in the range of $10^{-5}$.
        \item If $\hat{y}$ is taken as input to the inverted neural network, \textbf{the generated image does not meet expectations}.

    
        
\end{enumerate}

We found that if $y_{pred} + noise$, where $noise  \sim \mathcal{N}(0, 10^{-4})$ is passed as input to the inverted neural network, the input image $x$ is generated with acceptable quality. Our future work would be to perform training such that the error between $y_{pred}$ and $\hat{y}$ would be brought down to this range of $10^{-5}$ so that meaningful inversion happens.

%% file: ICVGIP-Latex-Template.bbl

\begin{thebibliography}{9}


\ifx \showCODEN    \undefined \def \showCODEN     #1{\unskip}     \fi
\ifx \showDOI      \undefined \def \showDOI       #1{#1}\fi
\ifx \showISBNx    \undefined \def \showISBNx     #1{\unskip}     \fi
\ifx \showISBNxiii \undefined \def \showISBNxiii  #1{\unskip}     \fi
\ifx \showISSN     \undefined \def \showISSN      #1{\unskip}     \fi
\ifx \showLCCN     \undefined \def \showLCCN      #1{\unskip}     \fi
\ifx \shownote     \undefined \def \shownote      #1{#1}          \fi
\ifx \showarticletitle \undefined \def \showarticletitle #1{#1}   \fi
\ifx \showURL      \undefined \def \showURL       {\relax}        \fi
\providecommand\bibfield[2]{#2}
\providecommand\bibinfo[2]{#2}
\providecommand\natexlab[1]{#1}
\providecommand\showeprint[2][]{arXiv:#2}

\bibitem[\protect\citeauthoryear{??}{dat}{aset}]%
        {dataset}
 \bibinfo{year}{Flickr Image dataset}\natexlab{}.
\newblock \bibinfo{booktitle}{\emph{2024}}.
\newblock
\urldef\tempurl%
\url{https://www.kaggle.com/datasets/hsankesara/flickr-image-dataset}
\showURL{%
\tempurl}


\bibitem[\protect\citeauthoryear{Bao, Chen, Wen, Li, and Hua}{Bao et~al\mbox{.}}{2017}]%
        {bao2017cvae}
\bibfield{author}{\bibinfo{person}{Jianmin Bao}, \bibinfo{person}{Dong Chen}, \bibinfo{person}{Fang Wen}, \bibinfo{person}{Houqiang Li}, {and} \bibinfo{person}{Gang Hua}.} \bibinfo{year}{2017}\natexlab{}.
\newblock \showarticletitle{CVAE-GAN: fine-grained image generation through asymmetric training}. In \bibinfo{booktitle}{\emph{Proceedings of the IEEE international conference on computer vision}}. \bibinfo{pages}{2745--2754}.
\newblock


\bibitem[\protect\citeauthoryear{Conneau, Kiela, Schwenk, Barrault, and Bordes}{Conneau et~al\mbox{.}}{2017}]%
        {conneau2017supervised}
\bibfield{author}{\bibinfo{person}{Alexis Conneau}, \bibinfo{person}{Douwe Kiela}, \bibinfo{person}{Holger Schwenk}, \bibinfo{person}{Lo{\"\i}c Barrault}, {and} \bibinfo{person}{Antoine Bordes}.} \bibinfo{year}{2017}\natexlab{}.
\newblock \showarticletitle{Supervised learning of universal sentence representations from natural language inference data}.
\newblock \bibinfo{journal}{\emph{arXiv preprint arXiv:1705.02364}} (\bibinfo{year}{2017}).
\newblock


\bibitem[\protect\citeauthoryear{Ho, Jain, and Abbeel}{Ho et~al\mbox{.}}{2020}]%
        {ho2020denoising}
\bibfield{author}{\bibinfo{person}{Jonathan Ho}, \bibinfo{person}{Ajay Jain}, {and} \bibinfo{person}{Pieter Abbeel}.} \bibinfo{year}{2020}\natexlab{}.
\newblock \showarticletitle{Denoising diffusion probabilistic models}.
\newblock \bibinfo{journal}{\emph{Advances in neural information processing systems}}  \bibinfo{volume}{33} (\bibinfo{year}{2020}), \bibinfo{pages}{6840--6851}.
\newblock


\bibitem[\protect\citeauthoryear{Ramesh, Dhariwal, Nichol, Chu, and Chen}{Ramesh et~al\mbox{.}}{2022}]%
        {ramesh2022hierarchical}
\bibfield{author}{\bibinfo{person}{Aditya Ramesh}, \bibinfo{person}{Prafulla Dhariwal}, \bibinfo{person}{Alex Nichol}, \bibinfo{person}{Casey Chu}, {and} \bibinfo{person}{Mark Chen}.} \bibinfo{year}{2022}\natexlab{}.
\newblock \showarticletitle{Hierarchical text-conditional image generation with clip latents}.
\newblock \bibinfo{journal}{\emph{arXiv preprint arXiv:2204.06125}} \bibinfo{volume}{1}, \bibinfo{number}{2} (\bibinfo{year}{2022}), \bibinfo{pages}{3}.
\newblock


\bibitem[\protect\citeauthoryear{Ramesh, Pavlov, Goh, Gray, Voss, Radford, Chen, and Sutskever}{Ramesh et~al\mbox{.}}{2021}]%
        {ramesh2021zero}
\bibfield{author}{\bibinfo{person}{Aditya Ramesh}, \bibinfo{person}{Mikhail Pavlov}, \bibinfo{person}{Gabriel Goh}, \bibinfo{person}{Scott Gray}, \bibinfo{person}{Chelsea Voss}, \bibinfo{person}{Alec Radford}, \bibinfo{person}{Mark Chen}, {and} \bibinfo{person}{Ilya Sutskever}.} \bibinfo{year}{2021}\natexlab{}.
\newblock \showarticletitle{Zero-shot text-to-image generation}. In \bibinfo{booktitle}{\emph{International conference on machine learning}}. Pmlr, \bibinfo{pages}{8821--8831}.
\newblock


\bibitem[\protect\citeauthoryear{Razavi, Van~den Oord, and Vinyals}{Razavi et~al\mbox{.}}{2019}]%
        {razavi2019generating}
\bibfield{author}{\bibinfo{person}{Ali Razavi}, \bibinfo{person}{Aaron Van~den Oord}, {and} \bibinfo{person}{Oriol Vinyals}.} \bibinfo{year}{2019}\natexlab{}.
\newblock \showarticletitle{Generating diverse high-fidelity images with vq-vae-2}.
\newblock \bibinfo{journal}{\emph{Advances in neural information processing systems}}  \bibinfo{volume}{32} (\bibinfo{year}{2019}).
\newblock


\bibitem[\protect\citeauthoryear{Reimers}{Reimers}{2019}]%
        {reimers2019sentence}
\bibfield{author}{\bibinfo{person}{N Reimers}.} \bibinfo{year}{2019}\natexlab{}.
\newblock \showarticletitle{Sentence-BERT: Sentence Embeddings using Siamese BERT-Networks}.
\newblock \bibinfo{journal}{\emph{arXiv preprint arXiv:1908.10084}} (\bibinfo{year}{2019}).
\newblock


\bibitem[\protect\citeauthoryear{S{\o}nderby, Poole, and Mnih}{S{\o}nderby et~al\mbox{.}}{2017}]%
        {sonderby2017continuous}
\bibfield{author}{\bibinfo{person}{Casper~Kaae S{\o}nderby}, \bibinfo{person}{Ben Poole}, {and} \bibinfo{person}{Andriy Mnih}.} \bibinfo{year}{2017}\natexlab{}.
\newblock \showarticletitle{Continuous relaxation training of discrete latent variable image models}. In \bibinfo{booktitle}{\emph{Beysian DeepLearning workshop, NIPS}}, Vol.~\bibinfo{volume}{201}.
\newblock


\end{thebibliography}
